\documentclass[lettersize,journal]{IEEEtran}

\usepackage{amsmath,amsfonts}
\usepackage{algorithmic}
\usepackage{algorithm}
\usepackage{array}
\usepackage[caption=false,font=normalsize,labelfont=sf,textfont=sf]{subfig}
\usepackage{textcomp}
\usepackage{stfloats}
\usepackage{url}
\usepackage{verbatim}
\usepackage{graphicx}
\usepackage{cite}
\RequirePackage{xspace}
\makeatletter
\DeclareRobustCommand\onedot{\futurelet\@let@token\@onedot}
\def\@onedot{\ifx\@let@token.\else.\null\fi\xspace}
\def\eg{\emph{e.g}\onedot} 
\def\ie{\emph{i.e}\onedot} 
 
\def\etc{\emph{etc}\onedot}

\def\etal{\emph{et al}\onedot}
\makeatother
\usepackage{multirow}
\usepackage{makecell}
\usepackage{cleveref}
\usepackage{xcolor}
\usepackage{booktabs}

\begin{document}

\title{AnimateAnywhere: Rouse the Background\\in Human Image Animation }


\author{Xiaoyu Liu,~Mingshuai Yao,~Yabo Zhang,~Xianhui Lin,~Peiran Ren,\\
        Xiaoming Li,~Ming Liu,~and~Wangmeng Zuo,~\IEEEmembership{Senior Member,~IEEE}

\thanks{Xiaoyu Liu, Mingshuai Yao, Yabo Zhang, Ming Liu, and Wangmeng Zuo are with the Faculty of Computing, Harbin Institute of Technology, Harbin, China (e-mail: liuxiaoyu1104@gmail.com; ymsoyosmy@gmail.com; hitzhangyabo2017@gmail.com; csmliu@outlook.com; wmzuo@hit.edu.cn).}%
\thanks{Xianhui Lin is an independent researcher (e-mail: xhlin129@gmail.com).}%
\thanks{Peiran Ren is an independent researcher (e-mail: renpeiran@gmail.com).}%
\thanks{Xiaoming Li is with Nanyang Technological University, Singapore (e-mail: csxmli@gmail.com).}%

}

\markboth{IEEE Transactions on Multimedia}%
{Liu \MakeLowercase{\textit{et al.}}: AnimateAnywhere: Rouse the Background in Human Image Animation}


\maketitle

\begin{abstract}
Human image animation aims to generate human videos of given characters and backgrounds that adhere to the desired pose sequences. However, existing methods focus more on human actions while neglecting the generation of background, which typically leads to static results or inharmonious movements.
The community has explored camera pose-guided animation tasks, yet preparing the camera trajectory is impractical for most entertainment applications and ordinary users.
As a remedy, we present an AnimateAnywhere framework, rousing the background in human image animation without requirements on camera trajectories.
In particular, based on our key insight that the movement of the human body often reflects the motion of the background, we introduce a background motion learner (BML) to learn background motions from human pose sequences.
To encourage the model to learn more accurate cross-frame correspondences of background, we further deploy an epipolar constraint on the 3D attention map. Specifically, the mask used to suppress geometrically unreasonable attention is carefully constructed by combining an epipolar mask and the current 3D attention map.
Extensive experiments demonstrate that our AnimateAnywhere effectively learns the background motion from human pose sequences, achieving state-of-the-art performance in generating human animation results with vivid and realistic backgrounds. Project page: \url{https://animateanywhere.github.io/}

\end{abstract}


\begin{figure}[t!]
    \centering
    \includegraphics[width=0.99\linewidth]{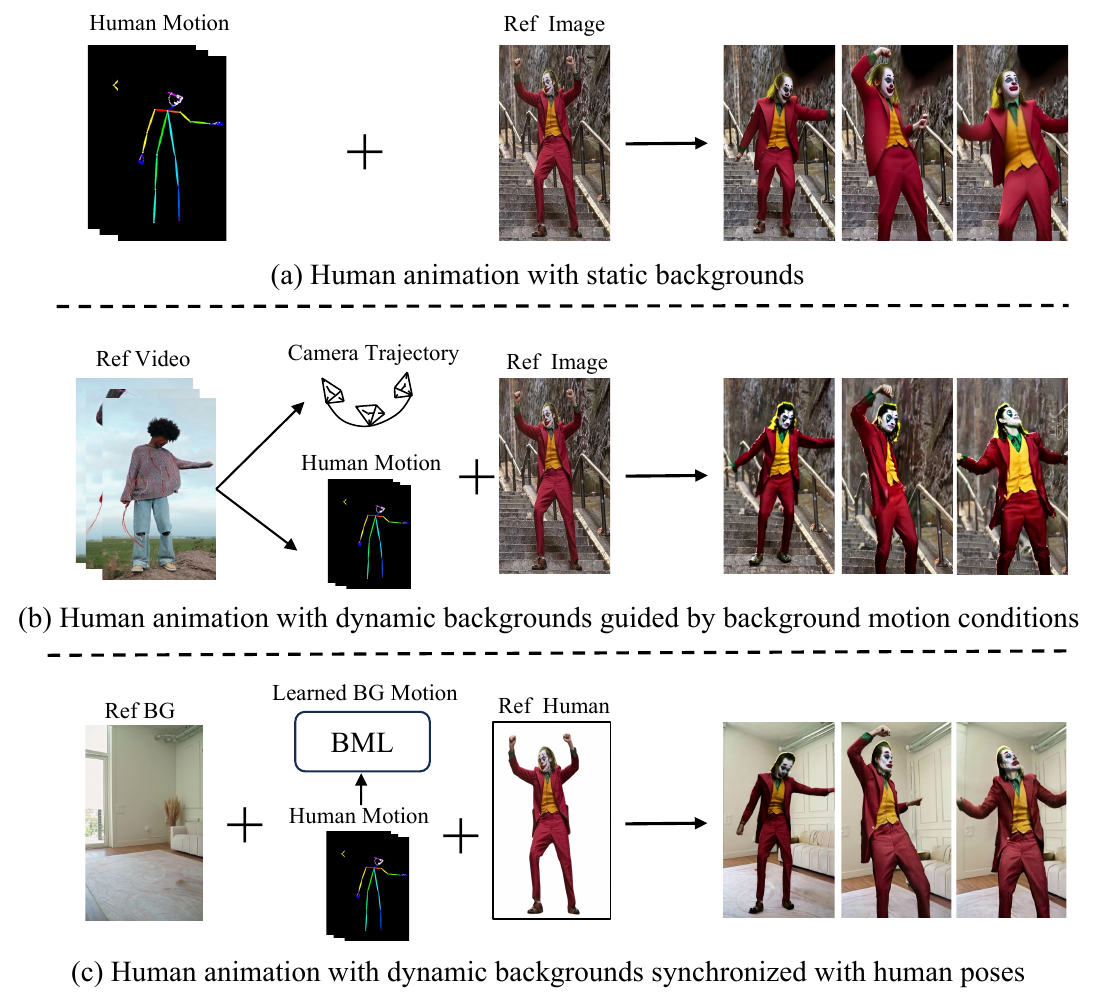}
    \vspace{-5mm}
    \caption{  Comparison with existing paradigms of human image animation. (a) Human animation with  static backgrounds: Animate Anyone~\cite{hu2024animate}, MagicAnimate~\cite{xu2024magicanimate}, Champ~\cite{zhu2024champ}, and \etc. They only generate static backgrounds.
   (b) Human animation with dynamic backgrounds guided by background motion conditions: Humanvid~\cite{wang2024humanvid}, Liu~\etal~\cite{liu2024disentangling}, and \etc. They extract background motion conditions of reference videos to guide the movement of the background. However, extracting background motion conditions paired with human poses heavily relies on reference videos, which are often unavailable in practice.
   (c) Ours: Human animation with dynamic backgrounds synchronized with human poses. Our paradigm learns the background movements that are harmonious with the human poses for human animation.}
    \label{fig:Animate_intro}
    \vspace{-6mm}
\end{figure}

\section{Introduction}

\IEEEPARstart{H}{uman} image animation has become a popular task focused on transforming source character images into realistic videos based on human pose sequences. 
This technique has a wide range of potential applications, including in entertainment videos, artistic creation, and virtual characters.

Recent advancements in generative diffusion models have significantly advanced the field of image animation~\cite{hu2024animate,xu2024magicanimate,zhu2024champ,zhou2024realisdance,wang2024realishuman,wang2024unianimate}, enabling the creation of highly realistic and controllable human image animation, particularly in terms of identity preservation and motion consistency.
Despite these advancements, most existing human image animation methods focus on improving the generation quality of foreground humans while leaving the background static or unconstrained.
However, in practice, the photographers usually need to adjust the cameras to better capture human movements, postures, and \etc, which results in background changes.
Thus, the static or disharmonious backgrounds are insufficient to meet the requirements for realistic and diverse human video generation.

To address this issue, one possible solution is to employ control signals to guide background motion.
For example, Humanvid~\cite{wang2024humanvid} controls the background motion via camera trajectories estimated from the reference videos, while Liu~\etal~\cite{liu2024disentangling} leverage sparse tracking points and human pose sequences to independently control background and foreground motions. 
However, these methods typically acquire camera trajectories and human pose sequences from a reference video, which is unavailable in many scenarios, such as science fiction film production.
Although camera trajectories can be manually designed, this easily results in disharmony between human and background motion. It requires professional skills and is redundant for most applications and ordinary users.
Additionally, another limitation of these methods is that a reference image with both the desired human (\eg, a clown) and desired background (\eg, the user’s living room) is required.

To overcome the above limitations of existing human image animation methods, this work proposes AnimateAnywhere, a novel framework that generates videos with harmonious human and background motion controlled solely by human pose sequences, without additional camera control conditions.
As shown in \cref{fig:Animate_intro}, the key idea of our AnimateAnywhere is to identify the background motion from human pose sequences.
We argue that human pose sequences can imply background motion. 
For instance, when a person appears larger or smaller across frames while remaining stationary, the background typically exhibits a corresponding zoom-in or zoom-out motion.
Moreover, if a person walks to the left while remaining centered in the frame, it means that the camera is moving left, and the background should shift accordingly.
By observing the relative changes in a person’s size and position across frames, as well as the type of human actions that can also be inferred from the human pose sequence, we can estimate both the direction and extent of background motion.

Specifically, we build the AnimateAnywhere framework upon a pre-trained image-to-video model (\ie, CogVideoX~\cite{yang2024cogvideox}) and independently inject a reference human image, a reference background image, and a human pose sequence into the generation process.
Thus, AnimateAnywhere supports free combinations of reference humans, backgrounds, and pose sequences, enabling highly flexible and controllable video generation.
To realize the key idea of AnimateAnywhere, we propose the Background Motion Learner (BML) module, which learns background motion from the latent features of human pose sequences by applying LoRA~\cite{hu2022lora} on the 3D-attention-based block.
Moreover, although 3D full attention allows each background pixel to attend to any other pixel from all frames, this highly free attention manner also brings difficulties to optimization, easily resulting in geometrically inconsistent background generation.
To further improve the cross-frame geometric consistency of the background, we propose an adaptive epipolar constraint for the 3D attention map. Therein, a mask used for suppressing geometrically unreasonable attention is elaborately designed from an epipolar mask and the current 3D attention map.

As shown in \cref{fig:Animate_intro}, our AnimateAnywhere can generate realistic videos with harmonious human and background motion, without additional background motion control conditions. 
Extensive experiments conducted on the constructed datasets show that our AnimateAnywhere outperforms state-of-the-art methods, demonstrating the effectiveness, flexibility, and controllability of generating dynamically harmonious human and background motions. Our contributions are listed as follows.

\begin{itemize}
    \item We propose AnimateAnywhere, a novel human image animation framework that generates dynamic backgrounds in harmony with human motions, requiring no camera pose information. It also supports flexible combinations of reference humans, backgrounds, and pose sequences.
    \vspace{1mm}
    \item We propose a Background Motion Learner module to predict background motion from human pose sequences and an adaptive epipolar constraint for the 3D attention map to enhance background geometric consistency.
    \vspace{1mm}
     \item Our method achieves state-of-the-art performance in the challenging test dataset, demonstrating the effectiveness, flexibility, and controllability for human animation.

\end{itemize}

\section{Related Work}

\subsection{Human Image Animation}

Human image Animation aims to generate coherent human videos from a reference image and a target pose sequence.
The GAN-based methods~\cite{siarohin2019first,wei2020gac,wang2021one,sarkar2021humangan,tian2021good,wang2022coherent,ji2025one} typically use deformation functions to spatially transform the reference image according to the input motion, thereby generating continuous video frames. 
However, these methods often suffer from unrealistic visual artifacts and temporal inconsistencies in the generated video.
By leveraging the generative visual capabilities of diffusion models, recent advancements~\cite{hu2024animate,xu2024magicanimate,zhu2024champ,zhou2024realisdance,wang2024realishuman,wang2024unianimate,wang2024humanvid,yao2025beyond} have enabled the generation of high-quality and consistent human animation videos. 
AnimateAnyone~\cite{hu2024animate} and MagicAnimate~\cite{xu2024magicanimate} introduce a video
diffusion model and a reference network for temporal consistency and identity preservation, respectively.
Champ~\cite{zhu2024champ} incorporates four distinct control signals to provide comprehensive 3D
shape and detailed pose guidance.
Following these, some works focus on realistic hands~\cite{zhou2024realisdance,wang2024realishuman}, efficient temporal modeling~\cite{wang2024unianimate}, or a pose alignment algorithm~\cite{musepose} to further improve the performance of human animation.
Animate-X~\cite{tan2024animate} extends animation to diverse character types with a unified framework.

Despite achieving reasonable results in maintaining temporal consistency in humans, these approaches face challenges in generating realistic and dynamic backgrounds.
MIMO~\cite{men2024mimo} and AnimateAnyone2~\cite{hu2025animate} address this challenge by taking the background video directly from a reference video as input.
Liu~\etal~\cite{liu2024disentangling} extract sparse point tracking from the reference video and use it to guide background animation.
Humanvid~\cite{wang2024humanvid} utilizes a SLAM-based method~\cite{wang2024tram, teed2021droid} to estimate an accurate camera trajectory from the original videos and leverages CameraCtrl~\cite{he2024cameractrl} and Animate Anyone~\cite{hu2024animate} to construct a human animation framework. DynaScene~\cite{yao2025beyond} further introduces a multi-task learning approach to enhance the realism of the generated background.
In comparison to the existing methods, our AnimateAnywhere predicts the background motion from human pose sequences without the reference video and camera pose extraction during inference.

\subsection{Camera Control Video Generation}

With the impressive capabilities of video diffusion models~\cite{ho2022video,guo2023animatediff,blattmann2023stable,bar2024lumiere,gupta2025photorealistic} in generating high-quality videos, several approaches have focused on achieving controllable video generation by incorporating conditional controls~\cite{zhang2023controlvideo,guo2025sparsectrl,wei2024dreamvideo,guo2023animatediff,he2024cameractrl,chen2025videodreamer}. In this section, we provide a brief overview of the progress in camera-controllable video generation.
AnimateDiff~\cite{guo2023animatediff} uses MotionLoRA~\cite{hu2021lora} fine-tuning to adapt a pre-trained video model to specific camera movements.
Direct-a-video~\cite{yang2024direct} introduces a camera embedder that incorporates three camera parameters, though the limited parameters restrict camera movement to translations and zooms. 
Following this, MotionCtrl~\cite{wang2024motionctrl} incorporates rotation and translation matrices derived from extrinsic camera parameters, enabling the generation of videos with more complex camera motions. CameraCtrl~\cite{he2024cameractrl} leverages Plücker coordinates to represent camera poses, allowing for geometric interpretations at the pixel level.
Recent advances further enhance the geometric consistency of generated videos by epipolar constraints or 3D reconstruction.
CamCo~\cite{xu2024camco} applies epipolar constraints to ensure consistency between generated frames and a single reference frame only.
CAMI2V~\cite{zheng2024cami2v} extends epipolar constraints onto all frames to account for scenarios where frames lack overlapping regions with the reference frame.
CamTrol~\cite{hou2024training} models camera movements through explicit 3D point clouds.
Building on the epipolar constraint used in CAMI2V~\cite{zheng2024cami2v}, we introduce an adaptive epipolar constraint on the 3D attention map of the background regions.

\begin{figure*}[t!]
    \centering
    \includegraphics[width=0.99\linewidth]{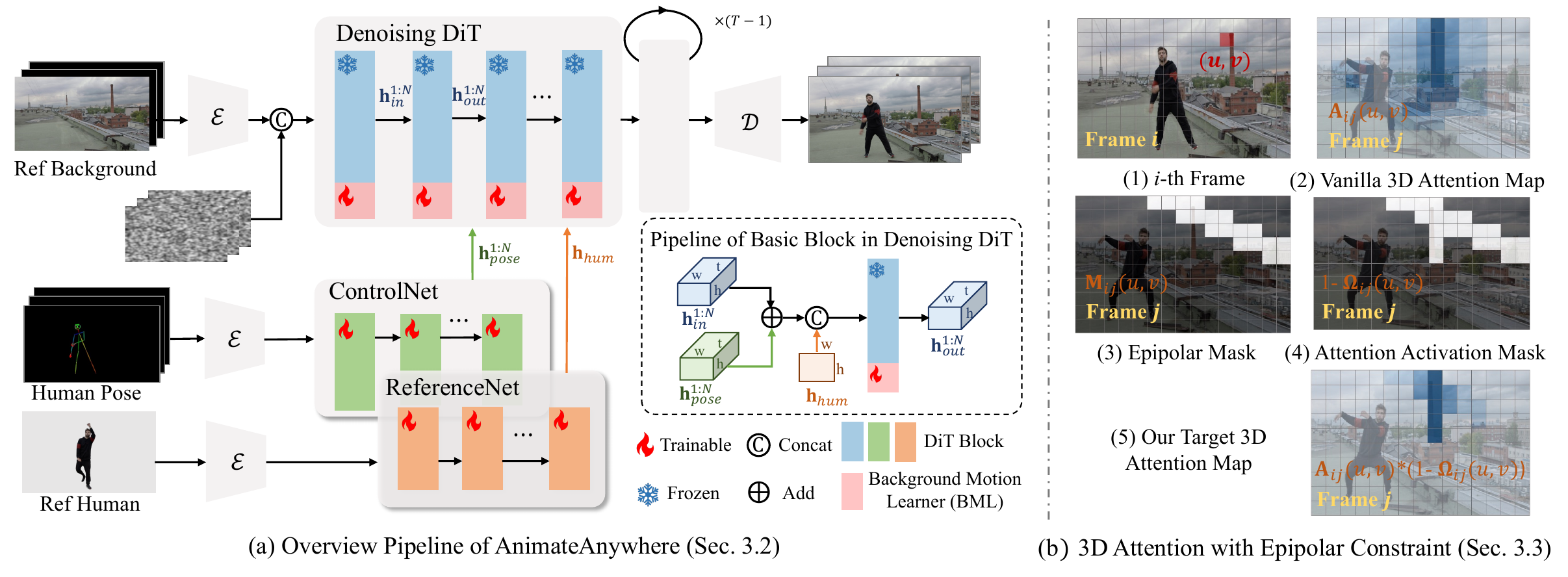}
    \vspace{-4mm}
    \caption{ (a) Overview Pipeline of AnimateAnywhere. Given a reference background image, a reference human image, and a human pose sequence as inputs, AnimateAnywhere generates photorealistic videos with synchronized motion of both the human and the background. Built upon CogVideoX~\cite{yang2024cogvideox}, we employ a ControlNet to inject human pose sequences and a ReferenceNet to incorporate the reference human. The background image is concatenated with noise to inject background appearance, while a Background Motion Learner (BML) predicts background motion from the pose sequence. (b) 3D Attention with Epipolar Constraint. (b)-(1) The $i$-th video frame.
(b)-(2) Vanilla 3D attention map $\mathbf{A}_{ij}(u,v)$ represents the model-learned attention weights from pixel $(u,v)$ in the $i$-th frame to all positions in the $j$-th frame.
(b)-(3) Epipolar mask $\mathbf{M}_{ij}(u,v)$ defines the geometrically valid region in the $j$-th frame.
(b)-(4) Attention activation mask $\mathbf{1} - \mathbf{\Omega}_{ij}(u,v)$ highlights the retained attention regions without constraint. Adaptive suppression mask $\mathbf{\Omega}_{ij}(u,v)$ selectively constrains low-confidence attention outside the epipolar region. (b)-(5) Target 3D attention map under our epipolar constraint. Our loss function is equivalent to computing the difference between $\mathbf{A}_{ij}(u,v)$ and this target map.
    }
    \vspace{-3mm}
    \label{fig:Animate_pipline}
\end{figure*}

\section{Method}

Given a human image $\mathbf{I}_\mathit{hum}$, a background image $\mathbf{I}_\mathit{bg}$, and a human pose sequence $\mathbf{P}^{1:N}$, AnimateAnywhere generates a video with synchronized human and background motion, guided solely by the pose sequence without explicit background motion control.
%
%
Firstly, the preliminaries of AnimateAnywhere, including our base model CogVideoX~\cite{yang2024cogvideox} and 3D Full Attention, are introduced in \cref{sec:Preliminaries}.
Additionally, we introduce the Background Motion Learner (BML) module that infers background motion representations from the latent features of the human pose sequence (as detailed in \cref{sec:animateanywhere}).
To further enhance the accuracy of background motion prediction, we incorporate epipolar masks derived from labeled camera trajectories as supervision (see  \cref{sec:epipolarmask}). Finally, the training objectives for AnimateAnywhere are described in \cref{sec:animateloss}.

\vspace{-2mm}
\subsection{Preliminaries}
\label{sec:Preliminaries}
\noindent\textbf{Image-to-video CogVideoX Model.} 
CogVideoX~\cite{yang2024cogvideox} is a video diffusion model with transformers (DiTs) that achieves high-quality video generation. 
Our AnimateAnywhere is built upon the Image-to-Video framework of CogVideoX~\cite{yang2024cogvideox}. 
Given a condition image, CogVideoX~\cite{yang2024cogvideox} first pads it with zeros to N frames. Then the padded input is encoded by the VAE and concatenated with noise. Subsequently, the combined latent is iteratively denoised over several steps to generate the final video.
CogVideoX~\cite{yang2024cogvideox} employs unified 3D full attention modules within each DiT block. 
The 3D full attention module performs calculations simultaneously in spatial and temporal dimensions, capturing both intra-frame and cross-frame interactions.
This enables CogVideoX~\cite{yang2024cogvideox} to comprehensively aggregate temporal information for video generation.

\noindent\textbf{3D Full Attention.} 
Firstly, the video latent $\mathbf{h} \in \mathbb{R}^{B \times F \times C \times H \times W}$ is reshaped as $\mathbf{h} \in \mathbb{R}^{B \times (H \times W \times F) \times C}$, where $F$, $C$, and $(H, W)$ denote the temporal, channel, and spatial dimensions, respectively. And then the reshaped latent is input into the 3D Full Attention module. This 3D Full Attention module achieves a better understanding of video content by unifying the modeling of both time and space. 
The attention map $\mathbf{A}$ is computed as:  
\begin{equation}
\mathbf{A} = \text{Softmax}\left(\frac{\mathbf{Q}(\mathbf{h}) \mathbf{K}(\mathbf{h})^\top}{\sqrt{d}}\right) \in \mathbb{R}^{B \times (H \times W \times F) \times (H \times W \times F)},
\end{equation}  
where $\mathbf{Q}$ and $\mathbf{K}$ denote the query and key projections, respectively. 
The resulting attention map $\mathbf{A}$ reflects the correspondences across video frames learned by the diffusion model.

\subsection{AnimateAnywhere}
\label{sec:animateanywhere}

As shown in \cref{fig:Animate_pipline}, AnimateAnywhere aims to simultaneously animate both the reference human and the reference background. The animation of the human is guided by the human pose sequence, while the background is guided by the implicit motion signals derived from the human pose sequence. Below, we describe the detailed process of human and
background animation.

\noindent\textbf{Human Animation.}
\textbf{(1) Reference Human Input.} 
To ensure that the human details are well preserved during animation, we introduce a ReferenceNet DiT, whose structure matches the first 18 blocks of the denoising DiT. 
Specifically, the reference human image is first encoded into latent space using a 3D VAE and subsequently passed through the ReferenceNet DiT to extract human features $\mathbf{h}_\mathit{hum}$.
The human features are concatenated with the features from the main denoising DiT. These combined features are then processed through 3D full attention layers while preserving the original dimensions of the denoising DiT.
\textbf{(2) Human Pose Sequence Guidance.}
We adopt an architecture similar to ControlNet~\cite{zhang2023adding,mou2024t2i} to guide human motion. Specifically, we replicate the first 18 blocks of the pre-trained denoising DiT to serve as the ControlNet DiT. We utilize the VAE encoder to transform the human pose sequence into latent features and then utilize the ControlNet DiT to extract multi-scale pose features $\mathbf{h}_{\text{pose}}^{1:N}$. Subsequently, these multi-scale features are added to the corresponding features of the denoising DiT via zero-initialized linear layers.

\noindent\textbf{Background Animation.}
\textbf{(1) Reference Background Input.}
Given a condition image, the CogVideoX~\cite{yang2024cogvideox} I2V model uses it as the first frame to generate a video. Following this setup, we treat the reference background as the condition image and the initial background of the generated video. The background image is initially padded to match the dimensions of the video, then encoded into latent features via the VAE encoder, and finally concatenated with noise to serve as input for the DiT module.
\textbf{(2) Background Motion Learner Module.}
Unlike other methods that extract camera trajectories from reference videos as background motion signals, we propose to directly predict background motion from human pose sequences. An intuitive approach is to explicitly learn camera trajectories from human pose sequences and incorporate the learned camera trajectories during video generation.
However, directly learning camera trajectories faces several challenges. 
First, splitting the task into two stages leads to error accumulation. Estimating camera trajectories in the form of RT matrices is inherently challenging, and any prediction error will be directly propagated and fixed in the second stage,  which is difficult to correct even with video generation priors.
Moreover, the labeled camera trajectories often suffer from scale inconsistency, which also increases the optimization difficulty.
In contrast, rather than explicitly decoupling camera trajectories, we directly utilize the background motion features embedded in human pose to drive background movement. In particular, we introduce a LoRA-based module named Background Motion Learner (BML) module into each DiT block, and this module follows the standard paradigm of LoRA~\cite{hu2022lora}, expressed as $\mathbf{W}_0 x  + \Delta \mathbf{W} x$. We add the human pose features $\mathbf{h}_\mathit{pose}$ and the denoising features $\mathbf{h}_\mathit{in}$ together to predict the background motion. The process is as follows:
\begin{equation} \mathbf{h}_\mathit{out}^{1:N} = \mathit{T} (\mathbf{h}_\mathit{in}^{1:N} + \mathbf{h}_\mathit{pose}^{1:N}) + \mathit{B}(\mathbf{h}_\mathit{in}^{1:N} + \mathbf{h}_\mathit{pose}^{1:N}),
\end{equation}
where $\mathit{T}$ and $\mathit{B}$ denote DiT block and BML module. Specifically, $\mathbf{h}_\mathit{in}$ has strong priors on the co-movement between humans and backgrounds. $\mathbf{h}_\mathit{pose}$, which is extracted from 3D attention in ControlNet, contains temporal information of human motion across frames. 
\begin{figure}[t!]
    \centering
    \includegraphics[width=0.99\linewidth]{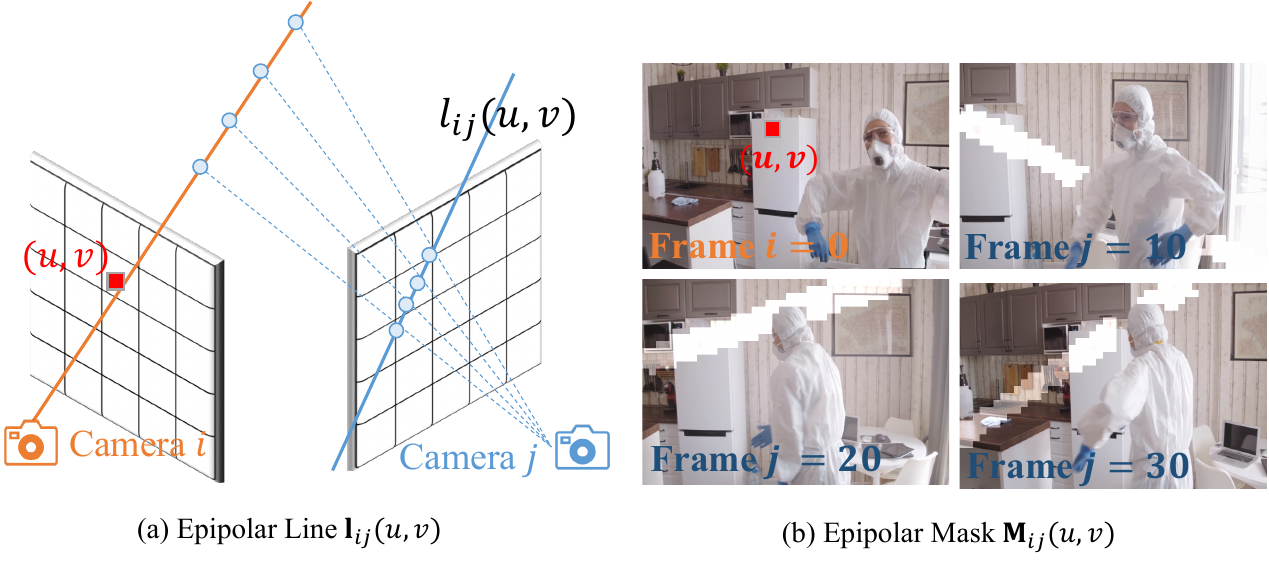}
    \vspace{-4mm}
    \caption{ Epipolar line and epipolar mask. (a) The epipolar line $\mathbf{l}_{ij}(u, v)$ is determined by the camera poses of the two frames. (b) The epipolar mask $\mathbf{M}_{ij}(u, v)$ is defined at the resolution of the latent feature space.}
    \label{fig:epipolar_line_mask}
    \vspace{-6mm}
\end{figure}

\subsection{ Epipolar Constraint on 3D Attention Map}
\label{sec:epipolarmask}
Although 3D full attention allows each background pixel to attend to any other pixel from all frames, this highly free search manner also makes optimization difficult and often causes geometric inconsistency in the background.
To address this, we introduce epipolar constraints into the 3D attention map. 
However, strictly restricting the correspondence search space to geometrically reasonable regions may undermine the cross-frame correlations learned by the original 3D full attention. So we propose an adaptive epipolar constraint that integrates an epipolar mask and the current 3D attention map. In the following, we first introduce the preliminaries of the epipolar constraint and the details of our proposed adaptive epipolar constraint on the 3D attention map.

\vspace{0.5em}
\noindent\textbf{Epipolar Line and Epipolar Mask.} 
The camera poses consist of intrinsic parameters $\mathbf{K} \in \mathbb{R}^{T \times 3 \times 3}$ and extrinsic parameters $[\mathbf{R};\mathbf{T}]$, where $\mathbf{R} \in \mathbb{R}^{T \times 3 \times 3}$ represents the rotational matrix and $\mathbf{T} \in \mathbb{R}^{T \times 3 \times 1}$ denotes the translational matrix.  
The epipolar constraint establishes that a point $(u, v)$ in the $i$-th frame must lie on its corresponding epipolar line $\mathbf{l}_\mathit{ij}(u, v)$ in the $j$-th frame, as illustrated in \cref{fig:epipolar_line_mask}. The epipolar line can be determined by camera poses as:  
\vspace{-1mm}
\begin{equation}
\mathbf{l}_\mathit{ij}(u, v) = \mathbf{F}_\mathit{ij} \cdot (u, v, 1)^T,
\end{equation} 
\vspace{-1mm}
where $\mathbf{F}_\mathit{ij}$ is the fundamental matrix between two frames. This matrix is computed as $ \mathbf{F}_\mathit{ij} = \mathbf{K}_\mathit{j}^{-T} \cdot (\mathbf{T}_\mathit{i \to j} \times \mathbf{R}_\mathit{i \to j}) \cdot \mathbf{K}_\mathit{i}^{-1}$. Here, $\mathbf{R}_\mathit{i \to j} \in \mathbb{R}^{3 \times 3}$ and $\mathbf{T}_\mathit{i \to j} \in \mathbb{R}^{3}$ are the relative rotation matrix and translation vector from $i$-th frame to $j$-th frame.  
Using the epipolar line, we convert each pixel $(u', v')$ in the $j$-th frame into an epipolar mask $\mathbf{M}_\mathit{ij}(u, v)$.  
The epipolar mask significantly reduces the number of matching points from $hw$ (frame size) to $l$ (where $l \ll hw$).  

\vspace{0.5em}
\noindent\textbf{Adaptive Epipolar Constraint.} 
We define the vanilla 3D attention map $\mathbf{A}_{ij}(u,v)$ as the attention intensity, indicating how much the pixel $(u,v)$ in the $i$-th frame attends to each position in the $j$-th frame. As described above, the epipolar mask $\mathbf{M}_{ij}(u,v)$ strictly defines the valid region in the $j$-th frame that the pixel $(u,v)$ in the $i$-th frame should correspond to under projective geometry.
To ensure that background motion adheres to geometric constraints, for each background pixel $(u,v)$ in $i$-th frame, we suppress the attention values of $\mathbf{A}_{ij}(u,v)$ that fall outside the corresponding epipolar region $\mathbf{M}_{ij}(u,v)$. 
However, completely suppressing attention outside the epipolar region $\mathbf{M}_{ij}(u,v)$ may disrupt the cross-frame correlations learned by 3D full attention. To mitigate this, we adaptively apply the epipolar constraint by incorporating a dynamic attention mask defined as (${\mathbf{A}_{ij}(u,v) < \delta}$), where (${\mathbf{A}_{ij}(u,v) < \delta}$) represents the region of low-confidence attention scores, \ie, the regions the network itself tends to ignore.
Thus, the suppressed regions are determined by the intersection:
$\mathbf{\Omega}_{ij}(u,v) = {(1-\mathbf{M}_{ij}(u,v))} \cap {(\mathbf{A}_{ij}(u,v) < \delta)}$. The final epipolar constraint is then defined by,
\begin{equation} \mathcal{L}_{\text{epipolar}} = \sum_{i,j} \sum_{(u,v)} \mathbf{\Omega}_{ij}(u,v)  \cdot \mathbf{A}_{ij}(u,v), \end{equation} 
where $\delta$ denotes a statistical threshold corresponding to the 30th percentile of attention values. Meanwhile, we emphasize that our loss function is equivalent to computing the difference between $\mathbf{A}_{ij}(u,v)$ and the target map, as illustrated in \cref{fig:Animate_pipline} (b)-(5).
During inference, our method no longer requires epipolar information or camera trajectories, and it directly leverages the attention learned through the epipolar loss.

\subsection{Learning Objective}
\label{sec:animateloss}
We employ the latent loss in CogVideoX~\cite{yang2024cogvideox} to train our pipeline,
\begin{equation}
    \mathcal{L}_\mathrm{latent} = \mathbb{E}_{\mathbf{z}_0^\mathit{1:N}, \mathbf{t}, \mathbf{c}, \epsilon}[\| \epsilon-\epsilon_\theta(\mathbf{z}_\mathit{t}^\mathit{1:N}, \mathbf{t},\mathbf{c} )) \|_2^2],
    \label{eqn:SmartControl_ldmloss}
\end{equation}
where $\epsilon_\theta$ denotes our diffusion model, $\mathbf{z}_0^{1:N}$ represents the latent embeddings of an $N$-frame image sequence, and $\mathbf{z}_t^{1:N}$ corresponds to the noisy latent at diffusion timestep $t$. The variable $\mathbf{c}$ denotes the conditioning inputs provided to our model, specifically including $\mathbf{I}_\mathit{hum}$, $\mathbf{I}_\mathit{bg}$, and $\mathbf{P}^\mathit{1:N}$, while $\epsilon$ denotes the unscaled noise.

Meanwhile, we employ a one-step sampling strategy to transform the predicted latent $\mathbf{z}_\mathit{t}^\mathit{1:N}$ at timestep $\mathbf{t}$ to $\mathbf{z}_\mathit{0}^\mathit{1:N}$ and we decode $\mathbf{z}_\mathit{0}^\mathit{1:N}$ into the RGB video $\mathbf{V}^\mathit{1:N}$ through the VAE decoder.
We employ the VGG perceptual loss between the generated frames $\hat{\mathbf{V}}^\mathit{1:N}$ and the target frames $\mathbf{V}^\mathit{1:N}$ during the last 30\% timesteps as follow:
\begin{equation}
    \mathcal{L}_\mathrm{vgg} = (1+ \mathbf{m})*VGG( \hat{\mathbf{V}}^\mathit{1:N}, \mathbf{V}^\mathit{1:N}),
\end{equation}
where $\mathbf{m}$ is the mask sequence of the human hand region.

The overall learning objective for training is defined by,
\begin{equation}
    \mathcal{L}=\mathcal{L}_\mathrm{latent} + \lambda_\mathrm{vgg}\mathcal{L}_\mathrm{vgg} + \lambda_\mathrm{epipolar}\mathcal{L}_\mathrm{epipolar},
\end{equation}
where $\lambda_\mathrm{vgg}$ and  $\lambda_\mathrm{epipolar}$ are the weights to balance loss terms.

\section{Experiments}

\subsection{Experimental Details}

\noindent\textbf{Dataset.}
We collect 5K videos from the Humanvid~\cite{wang2024humanvid} dataset and 3K videos from the Internet for training, where the Humanvid~\cite{wang2024humanvid} dataset has the annotation of camera pose sequence.
Our training dataset consists of videos captured under diverse camera conditions, including both static and moving cameras.

We evaluate our model on the Humanvid~\cite{wang2024humanvid} test dataset and a wild dataset, Bilibili200 (BL200), collected from the internet. The Humanvid test dataset includes 100 videos with noticeable background motion but limited human motion. The BL200 dataset contains 200 videos with significant motion in both the human and background.
For fair comparison, we evaluate all methods by generating 48-frame sequences with a stride of 4.
Notably, while other methods utilize a single reference image that contains both the human and background during testing, our method inputs the reference human and reference background separately.

\vspace{0.5em}
\noindent\textbf{Evaluation Metrics.}
Following the metrics utilized in~\cite{hu2024animate,xu2024magicanimate,zhu2024champ,zhou2024realisdance,wang2024realishuman,wang2024unianimate}, we use PSNR~\cite{huynh2008scope}, SSIM~\cite{wang2004image}, and LPIPS~\cite{zhang2018unreasonable} metrics to evaluate the quality of single-frame images. Additionally, we introduce FID~\cite{heusel2017gans} and FVD~\cite{unterthiner2019fvd} metrics to evaluate the difference between the distributions of generated videos and real-world videos.

\vspace{0.5em}
\noindent\textbf{Implementation Details.}
We adopt the pretrained CogVideoX1.5-5B-I2V~\cite{yang2024cogvideox} as our base model and keep it frozen throughout the training process. Both ControlNet and ReferenceNet are initialized with the parameters of CogVideoX~\cite{yang2024cogvideox}. Our training consists of three stages. In the first stage, we incorporate ControlNet into the base model and train only ControlNet while keeping all other parameters fixed. In the second stage, we introduce ReferenceNet and train it exclusively. In the final stage, we jointly optimize ControlNet, ReferenceNet, and the BML module.

In terms of data preparation, we first employ Grounded-SAM~\cite{ren2024grounded} to extract mask sequences in the human body and hand regions for each video. Subsequently, we obtain the human sequence and the background sequence using the extracted mask sequence. We randomly select a frame in the human sequence as the reference human and take the first frame from the background sequence as the reference background. Specifically, we complete the reference background using the inpainting method~\cite{zhou2023propainter}.
All experiments are trained on a mixture of horizontal (1360$\times$768) and vertical (768$\times$1360) videos and conducted on two A100 GPUs.
Training is performed over 20,000 steps with a batch size of 8. We employ the AdamW optimizer~\cite{loshchilov2017decoupled} with a constant learning rate of $1 \times 10^{-5}$.
We set the trade-off weights $\lambda_\mathrm{vgg}$ and $\lambda_\mathrm{epipolar}$ to 0.2 and 0.005, respectively. During inference, classifier-free guidance (CFG) is applied with a scale of 2.5 for both reference images and pose sequences.

\begin{table*}
\centering
\caption{Quantitative comparison for human animate on Humanvid~\cite{wang2024humanvid} test dataset and BL200 dataset. The best results are highlighted with \textbf{bold}. Results marked by $\dagger$ are retrained with the Humanvid~\cite{wang2024humanvid} train dataset, respectively.
Results marked with ${*}$ indicate that camera poses are required during testing.
Since the BL200 test dataset lacks labeled camera poses, the results of Humanvid$^\ddagger$~\cite{wang2024humanvid} are not available.}
\vspace{-3mm}
\resizebox{\textwidth}{!}{
\begin{tabular}{c|ccccc|ccccc} 
\toprule

\multirow{2}{*}{Methods} &\multicolumn{5}{c|}{Humanvid} &\multicolumn{5}{c}{BL200}\\
\cmidrule(lr){2-11}
  & PSNR$\uparrow$  & SSIM$\uparrow$ &LPIPS$\downarrow$  & FID$\downarrow$  & FVD$\downarrow$  &  PSNR$\uparrow$  & SSIM$\uparrow$ &LPIPS$\downarrow$  & FID$\downarrow$  & FVD$\downarrow$ \\
\midrule
MagicAnimate
~\cite{xu2024magicanimate} &  16.01 & 0.5463 & 0.4455 & 78.01 & 1943.07 &  16.82 & 0.5914 & 0.4133 & 40.02 & 1884.48  \\
AnimateAnyone
~\cite{hu2024animate}   & 16.47 &  0.5311 & 0.3913 & 50.18 & 1431.43 & 17.58 &  0.5756 & 0.3635 & 24.46 & 1507.05 \\
Champ~\cite{zhu2024champ}  & 15.08 & 0.5075 & 0.4481 & 63.92 & 1837.07 & 16.88 &  0.5649 & 0.3919 & 30.69 & 1701.45  \\
MusePose~\cite{musepose}  & 16.93 & 0.5578 &0.3676 & 52.00 & 1784.85 & 18.67 & 0.6126 & 0.3220 &22.86 & 1580.47 \\
DynamiCtrl~\cite{zhao2025dynamictrl}	& 15.15 &	0.5054 & 0.4032 & 46.16 & 1907.71& 16.22 & 0.5753  & 0.3511 & 26.75 &2025.79  \\
AnimateAnyone$^{\dagger}$~\cite{hu2024animate} &16.81 & 0.5486 &0.3672 & 47.69 & 1235.65 & 18.23 &  0.5930 & 0.3379 & 23.27 & 1371.22 \\
MusePose$^{\dagger}$~\cite{musepose}  & 17.33 & 0.5629 & 0.3563 & 44.73 & 1362.14 & 18.60 & 0.6116 & 0.3245 & 22.09 & 1444.79\\
DynamiCtrl$^\dagger$~\cite{zhao2025dynamictrl}   & 17.27 &0.5694 &0.3462 &44.84 &	1205.63 & 18.19 & 0.5901  & 0.3340 & 22.70 &1565.75  \\
Humanvid$^{*}$~\cite{wang2024humanvid}  & 17.91 &0.5774 & \textbf{0.3007} &46.66 &	991.66 & - & - & - & - & - \\
\midrule
Ours & \textbf{18.07} & \textbf{0.5831} & 0.3119 & \textbf{44.51} & \textbf{978.09} & \textbf{18.99} & \textbf{0.6182} & \textbf{0.3206} & \textbf{21.31} & \textbf{1154.27}\\
\bottomrule
\end{tabular}
\label{tab:comparisonanimate}}
\end{table*}

\begin{figure*}[t!]
    \centering
     \includegraphics[width=0.96\linewidth]{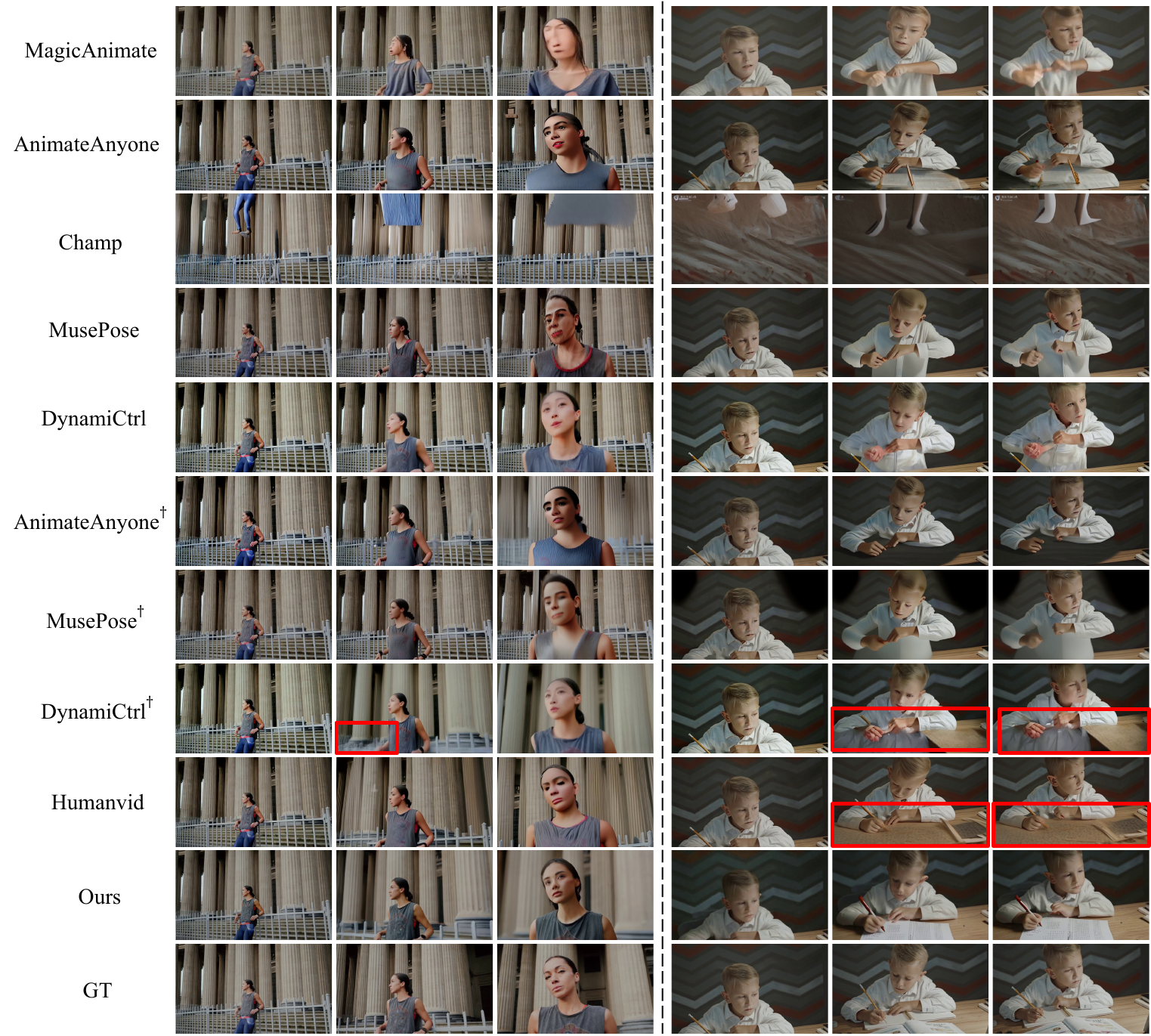}
    \vspace{-4mm}
    \caption{ Qualitative comparison on Humanvid~\cite{wang2024humanvid} test dataset. Our proposed method generates more realistic videos while ensuring harmonious motion between the foreground and background. Champ~\cite{zhu2024champ} predicts SMPL parameters, including camera parameters, but when the camera moves significantly, the estimated camera parameters become unreliable.  }
    \vspace{-3mm}
    \label{fig:Animate_compare}
\end{figure*}

\begin{figure*}[t!]
    \centering
     \includegraphics[width=0.96\linewidth]{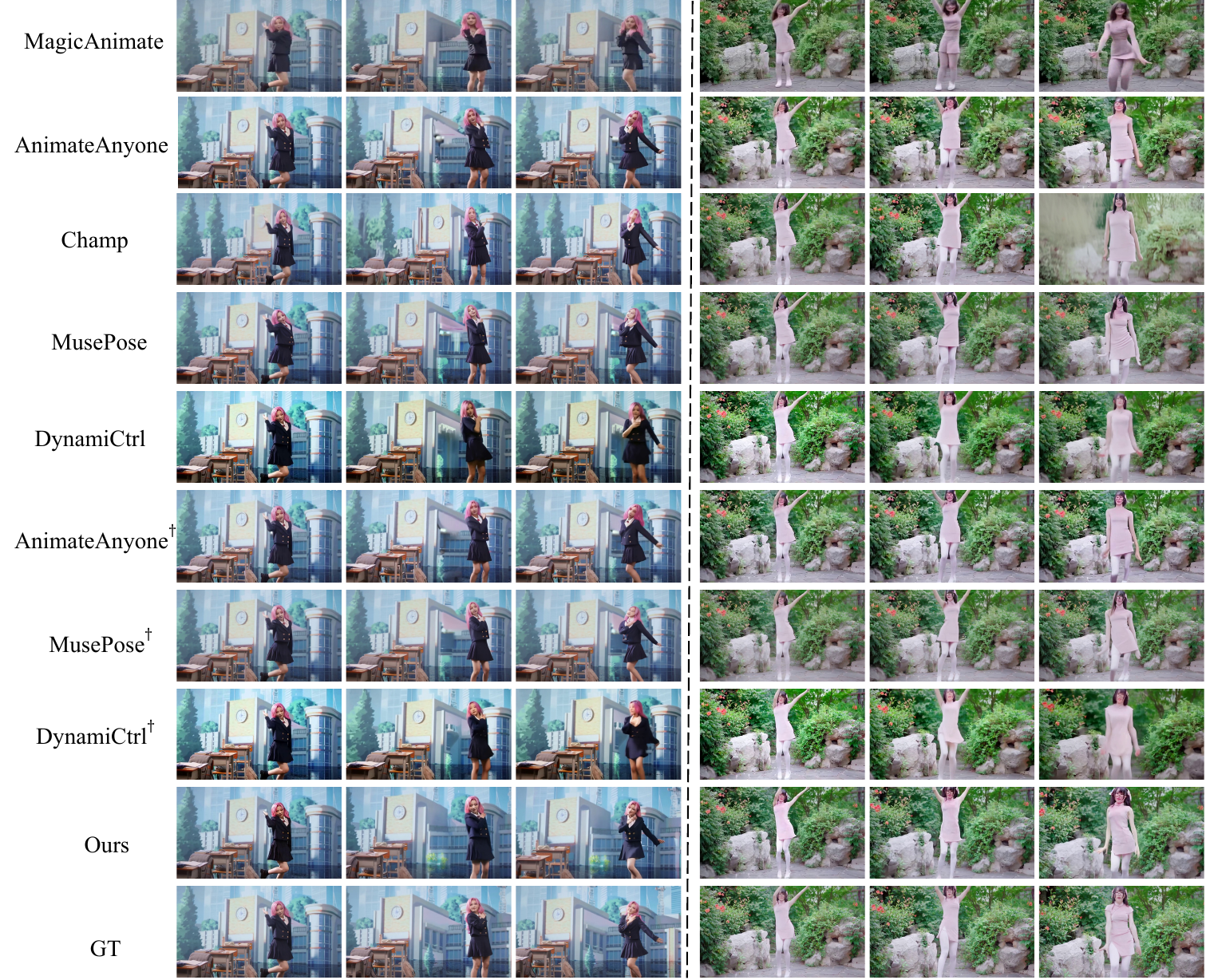}
    \vspace{-3mm}
    \caption{
Qualitative comparisons on the BL200 test dataset. Our method not only generates realistic background motion but also maintains the temporal consistency of both the human and background.}
    \label{fig:Animate_compare_2}
    \vspace{-3mm}
\end{figure*}

\subsection{Comparison with Existing Methods}

We compare our method with existing state-of-the-art human animation methods in both quantitative and qualitative aspects, including MagicAnimate~\cite{xu2024magicanimate}, AnimateAnyone~\cite{hu2024animate}, Champ~\cite{zhu2024champ}, MusePose~\cite{musepose}, DynamiCtrl~\cite{zhao2025dynamictrl}, and Humanvid~\cite{wang2024humanvid}.
Since AnimateAnyone~\cite{hu2024animate} is not officially open-sourced, we adopt its publicly available reimplementation from Moore-AnimateAnyone\footnote{https://github.com/MooreThreads/Moore-AnimateAnyone} for evaluation.
Note that these methods, except Humanvid~\cite{wang2024humanvid}, are originally trained on datasets with static backgrounds.
For fair comparison, we retrain AnimateAnyone~\cite{hu2024animate} MusePose~\cite{musepose} and DynamiCtrl~\cite{zhao2025dynamictrl} on our training dataset. 
Specifically, AnimateAnyone~\cite{hu2024animate} and MusePose~\cite{musepose} are built upon the AnimateDiff~\cite{guo2023animatediff}, while DynamiCtrl~\cite{zhao2025dynamictrl} adopts CogVideoX~\cite{yang2024cogvideox} as base models.

\vspace{0.5em}

\noindent\textbf{Quantitative Comparison.}
We conduct comprehensive experiments to assess the effectiveness of the proposed method, and the quantitative results are shown in \cref{tab:comparisonanimate}.
From all metrics on the Humanvid~\cite{wang2024humanvid} test dataset and the BL200 test dataset, we can see that previous methods originally struggled to generate videos with dynamic backgrounds, resulting in significantly lower metrics.
Even the retrained versions of AnimateAnyone~\cite{hu2024animate}, MusePose~\cite{musepose}, and DynamiCtrl~\cite{zhao2025dynamictrl} still yield inferior results as they still exhibit static or inconsistent backgrounds.
Humanvid~\cite{wang2024humanvid} achieves reasonable performance by relying on precisely estimated camera poses. 
In contrast, our method does not depend on camera pose and instead leverages the BML module together with the epipolar loss to effectively model dynamic backgrounds. 
As a result, our approach not only exhibits significant improvement over baseline methods but also surpasses Humanvid~\cite{wang2024humanvid} on most metrics.

\vspace{0.5em}
\noindent\textbf{Qualitative Comparison}.
The qualitative comparison with competing methods is shown in \cref{fig:Animate_compare}.
MagicAnimate~\cite{xu2024magicanimate}, AnimateAnyone~\cite{hu2024animate}, Champ~\cite{zhu2024champ}, MusePose~\cite{musepose}, and DynamiCtrl~\cite{zhao2025dynamictrl}  generate videos with static backgrounds that are usually unrealistic. 
Even after retraining, AnimateAnyone~\cite{hu2024animate} and MusePose~\cite{musepose} show only marginal improvements and still fail to generate dynamic backgrounds in many cases. For example, the left case in \cref{fig:Animate_compare} exhibits a completely static background, while the right case yields an inconsistent scene where only the wall moves, but the table remains still.
Although retrained DynamiCtrl~\cite{zhu2024champ} and Humanvid are more capable of dynamic backgrounds, the generation regions are often inconsistent with the reference background or contain noticeable artifacts, which are highlighted by red boxes in \cref{fig:Animate_compare}.
In contrast, our method synthesizes photorealistic videos while maintaining coherent motion between the human and the background. We also provide more generated videos on the BL200 dataset in \cref{fig:Animate_compare_2}. These results also highlight our method's capability to generate realistic background motion while maintaining high-quality, temporally consistent animation of both the human and the background.
Moreover, our method achieves superior identity preservation, particularly in face regions, compared with other methods.
We provide videos of these examples and more examples on the project page.

\begin{figure*}[t!]
    \centering
    \includegraphics[width=0.99\linewidth]{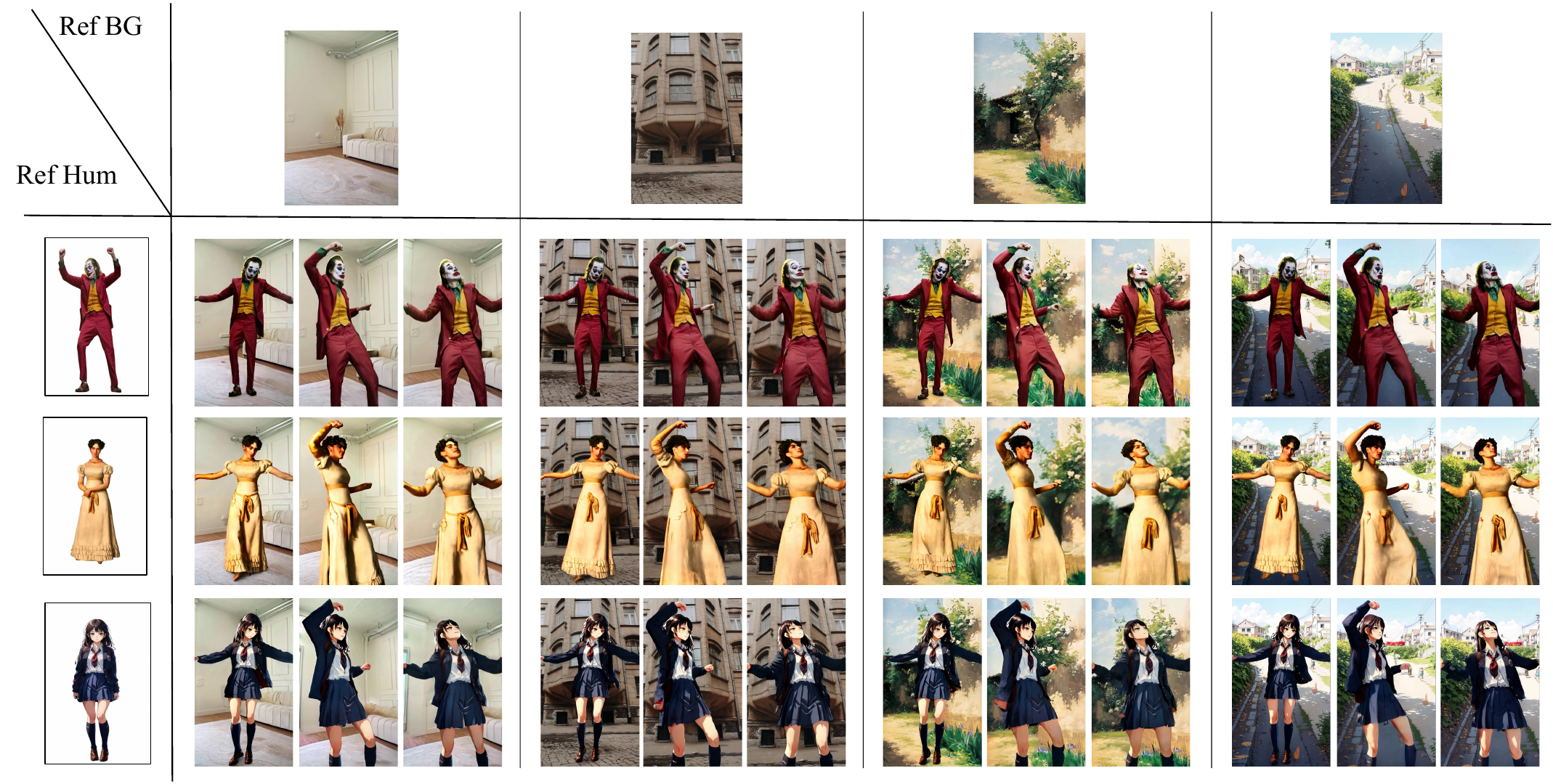}
    \vspace{-4mm}
    \caption{
   \textbf{The results are obtained from our method by cross-using the reference human image and background image.} 
    }
    \label{fig:Animate_cross}
    \vspace{-2mm}
\end{figure*}

\begin{table}
\centering
\small
\caption{ \textbf{Ablation study on the  epipolar loss $\mathcal{L}_\mathrm{epipolar}$.}}
\vspace{-2mm}
\resizebox{0.5\textwidth}{!}{
\begin{tabular}{cccccc}
\toprule
 Method& PSNR$\uparrow$  & SSIM$\uparrow$ &LPIPS$\downarrow$  & FID$\downarrow$  & FVD$\downarrow$  \\
\midrule
Ours w/o $\mathcal{L}_{epipolar}$ & 15.83 & 0.4823 & 0.4063 & 96.48 & 1269.42\\
Ours with $\mathcal{L}_{epipolar}$ ($\mathbf{1}-\mathbf{M}$) & 14.71 & 0.4315 & 0.4261 & 114.26 & 1373.42 \\
Ours with $\mathcal{L}_{epipolar}$ ($(\mathbf{1}-\mathbf{M}) \cap (\mathbf{A} < \delta$)) & \textbf{16.01} & \textbf{0.5831} & \textbf{0.3886} & \textbf{91.29} & \textbf{1243.07}\\
\bottomrule
\end{tabular}
\label{tab:Animate_epipolar}}
\vspace{-1mm}
\end{table}

\begin{figure}[t!]
    \centering
    \includegraphics[width=0.99\linewidth]{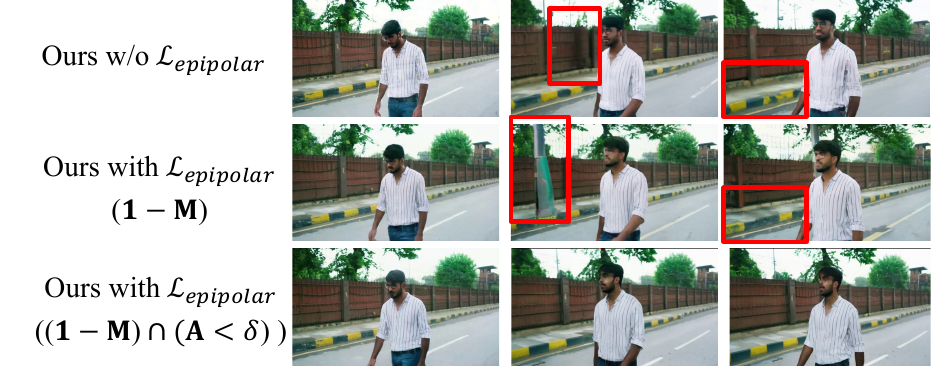}
    \vspace{-5mm}
    \caption{ Qualitative comparison of epipolar loss. Without the epipolar loss or with ${(\mathbf{1} -\mathbf{M})}$, the background contains obvious artifacts highlighted by red boxes. In contrast, our proposed epipolar loss yields consistent backgrounds.
    }
    \label{fig:Animate_epipolar_ab}
    \vspace{-4mm}
\end{figure}

\vspace{0.5em}
\noindent\textbf{Animation Cross Human and Background.}
Existing methods typically rely on a single reference image that includes both the human and the background, which limits the ability to customize videos for individual characters and backgrounds. In contrast, our method allows for separate input of the reference human and the reference background,  enhancing the flexibility of the generation process. To validate the effectiveness of our approach, we present additional results of different characters and backgrounds in \cref{{fig:Animate_cross}}, further confirming the unique advantages of our method.

\subsection{Ablation Study}

\vspace{0.5em}

\begin{table}
\centering
\small
\caption{\textbf{Ablation study on the strategies for background motion learning.}}
\vspace{-2mm}
\scalebox{0.85}{
\begin{tabular}{c|ccccc}
\toprule
 Method & PSNR$\uparrow$ & SSIM$\uparrow$ & LPIPS$\downarrow$ & FID$\downarrow$ & FVD$\downarrow$ \\
\midrule
 Explicit Two-stage & 17.00 & 0.5646 &0.3841 & 61.07 & 1604.10 \\
 Ours w/o BML & 16.86 & 0.5676 & 0.3333 & 51.15 & 1199.54 \\
 Ours & \textbf{18.07} & \textbf{0.5831} & \textbf{0.3119} & \textbf{44.51} & \textbf{978.09} \\
\bottomrule
\end{tabular}
}
\label{tab:bml_ablation}
\vspace{-4mm}
\end{table}

\noindent\textbf{Effects of Background Motion Learner Module}.
As illustrated in \cref{sec:animateanywhere}, we introduce a Background Motion Learner (BML) module that directly infers background motion from human pose sequences. In this subsection, we evaluate three strategies for background motion learning: (1) an explicit two-stage approach, (2) without BML, and (3) the proposed BML module. As shown in \cref{tab:bml_ablation}, the two-stage method suffers from error accumulation and scale inconsistency in labeled trajectories, leading to degraded performance. In contrast, our BML reformulates the task as an end-to-end background motion learning problem, achieving the best overall results. Furthermore, compared with the model without BML, our method improves PSNR by 7.2\% and FID by 13.0\%, clearly validating the effectiveness of the proposed module.

\vspace{0.5em}
\noindent\textbf{Effects of Epipolar Loss}.
In \cref{sec:epipolarmask}, we designed the epipolar loss to better learn background motion. However, some videos in Humanvid~\cite{wang2024humanvid} test dataset contain static or weak background motion, which diminishes its effect. To better assess the effects of epipolar loss, we select a subset with significant motion based on camera parameters $\mathbf{R}$ and $\mathbf{T}$ of the start and end frames:
$\left\| \mathbf{T}_{\text{end}} - \mathbf{T}_{\text{start}} \right\| > \tau_t 
\quad \lor \quad 
\arccos\left(\frac{\mathrm{Tr}(\mathbf{R}_{\text{end}} \mathbf{R}_{\text{start}}^\top) - 1}{2}\right) > \tau_r,$
where $\tau_t$ and $\tau_r$ are the 10th largest translation and rotation differences in the test set.
As shown in \cref{tab:Animate_epipolar}, our method with $\mathcal{L}_{\mathrm{epipolar}}$ $(\mathbf{1}-\mathbf{M}) \cap (\mathbf{A} < \delta)$ outperforms both the one without epipolar loss and the one using the mask $(1-\mathbf{M})$, demonstrating its effectiveness.
Qualitative comparisons in \cref{fig:Animate_epipolar_ab} further illustrate this.
Without epipolar loss, the generated backgrounds exhibit noticeable artifacts as the learned background motion does not adhere to geometric constraints. Additionally, suppressing all attention values outside the epipolar mask does not alleviate this issue but performs even worse. As evidenced in the second row of \cref{fig:Animate_epipolar_ab}, noticeable artifacts and unexpected objects appear in the background. In contrast, our approach achieves desired results by applying $\mathcal{L}_\mathrm{epipolar}$ to the region of $(\mathbf{1}-\mathbf{M}) \cap (\mathbf{A} < \delta)$. 

\begin{figure}[t!]
    \centering
    \includegraphics[width=0.99\linewidth]{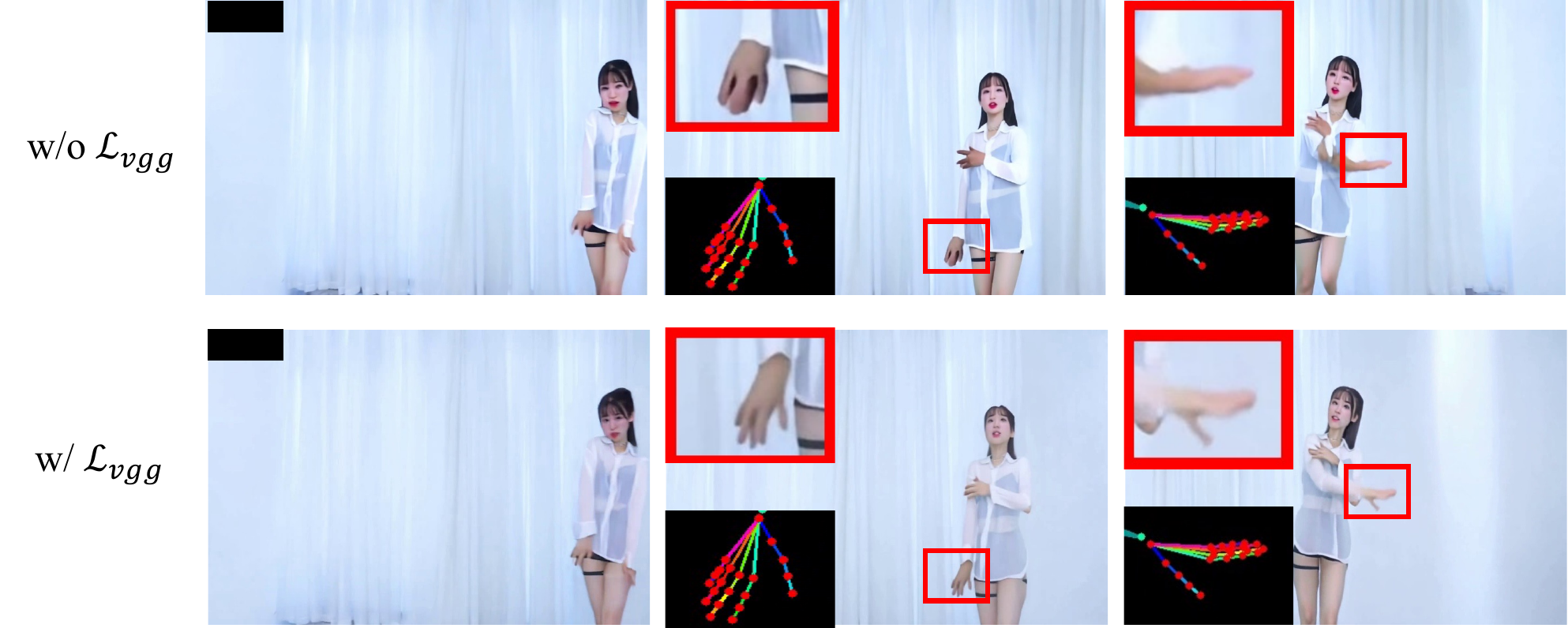}
    \vspace{-5mm}
    \caption{The generated results with and without the VGG perceptual loss. The result with VGG perceptual loss better preserves the fine details highlighted by the red box.}
    \label{fig:Animate_vgg}
    \vspace{-5mm}
\end{figure}

\vspace{0.5em}

\noindent\textbf{Effects of VGG Loss}.
We investigate the impact of the VGG perceptual loss $\mathcal{L}_\mathrm{vgg}$ on video generation. While the 3D VAE used in CogVideoX~\cite{yang2024cogvideox} achieves efficient spatiotemporal compression (8$\times$ spatially, 4$\times$ temporally), we observe that optimizing solely with an L2 loss in latent space leads to degraded high-frequency details. As shown in \cref{fig:Animate_vgg}, the hand regions exhibit blurred contours without $\mathcal{L}_\mathrm{vgg}$. After incorporating $\mathcal{L}_\mathrm{vgg}$ in pixel space, the details of the hand regions are well-preserved.

\vspace{0.5em}
\begin{table}
\centering
\small
\caption{\textbf{Performance under different values of the control parameter.}}
\vspace{-2mm}
\scalebox{0.85}{
\begin{tabular}{c|ccccc}
\toprule
Value & PSNR$\uparrow$ & SSIM$\uparrow$ & LPIPS$\downarrow$ & FID$\downarrow$ & FVD$\downarrow$ \\
\midrule
10 & 18.05 & 0.5804 & 0.3131 & 45.64 & 984.23 \\
30 & \textbf{18.07} & \textbf{0.5831} & \textbf{0.3119} & \textbf{44.51} & \textbf{978.09} \\
50 & 17.02 & 0.5671 & 0.3217 & 46.72 & 1087.21 \\
70 & 16.87 & 0.5571 & 0.3375 & 47.77 & 1229.87 \\
90 & 16.85 & 0.5479 & 0.3455 & 50.51 & 1289.06 \\
\bottomrule
\end{tabular}
}
\label{tab:parameter_effect}
\vspace{-4mm}
\end{table}

\noindent\textbf{Effects of threshold $\delta$}.
In our adaptive epipolar constraint, the threshold $\delta$ is used to identify low-confidence attention regions. 
We perform a sensitivity analysis by varying $\delta$ from the 0th to the 100th percentile, as shown in \cref{tab:parameter_effect}. 
When the threshold is too small ($\leq 10\%$), the epipolar constraint is inactive. 
In contrast, a high threshold ($\geq 50\%$) may interfere with the learned 3D attention, leading to degraded performance. We observe that setting $\delta$ to the 30th percentile yields the best performance and thus adopt this value.

\begin{figure}[t!]
    \centering
    \includegraphics[width=0.90\linewidth]{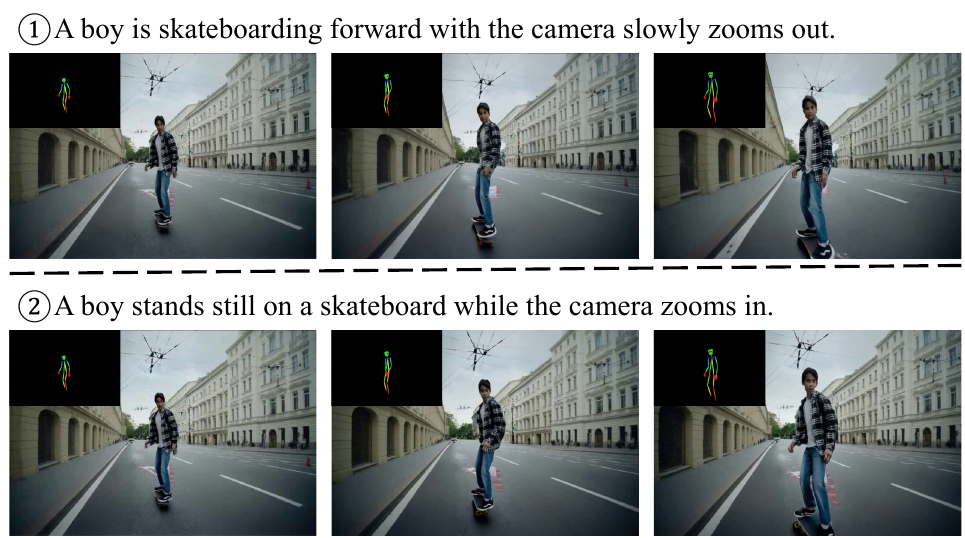}
    \vspace{-3mm}
    \caption{Examples of multiple camera poses based on a human pose sequence.}
    \label{fig:Animate_limit}
    \vspace{-5mm}
\end{figure}

\vspace{0.5em}

\section{Limitations.} 
Although AnimateAnywhere performs well in generating human animations with vivid and realistic backgrounds, it encounters challenges in predicting background motion under certain ambiguous conditions.
In such cases, a single human pose sequence may correspond to multiple plausible background motions. For example, considering a boy standing on a skateboard and appearing to grow larger, it is unclear whether the boy is skateboarding forward with the camera slowly zooming out (the boy’s forward motion is faster than the camera’s zoom-out) or the boy is standing still with the camera zooming in. This ambiguity arises because the action of skateboarding is difficult to infer solely from the human pose sequence.
Nevertheless, regardless of the above scenario, our method generates videos with plausible background motion as shown in \cref{fig:Animate_limit}. We provide corresponding videos on the project page.
To address this issue in the future, we plan to incorporate text descriptions that explicitly specify the intended background motion. This can help reduce ambiguity and better align the generated video with user expectations. 

\section{Conclusion}
\label{sec:conclusion}

In this paper, we introduce AnimateAnywhere, a human animation framework that generates videos with synchronized human and background motion without requiring specific background motion controls. 
The core of our framework is learning background motion representations directly from human pose sequences. 
Our method also provides additional geometric constraints on background motion
learning by leveraging the labeled camera poses during training.
Extensive experiments across multiple benchmarks validate the effectiveness of our AnimateAnywhere, demonstrating state-of-the-art performance in creating dynamic, realistic backgrounds that harmonize with human motion. 
This innovation not only improves the realism of the generated videos but also opens new possibilities in applications such as entertainment, virtual reality, and artistic content creation, where dynamic and immersive human-centered animations are increasingly valuable.


 
 \bibliographystyle{IEEEtran}
 \bibliography{references}


 




\vfill

\end{document}